\documentclass[letterpaper, 10 pt, conference]{ieeeconf}  

\IEEEoverridecommandlockouts                              
\usepackage{cite}
\usepackage{amsmath,amssymb,amsfonts}
\usepackage{algorithmic}
\usepackage{graphicx}
\usepackage{textcomp}
\usepackage{xcolor}
\usepackage{multirow}
\usepackage{booktabs}
\usepackage[hyphens]{url}
\usepackage{xcolor}

\title{\LARGE \bf
ZONDA: Zero-shot Object Navigation with Dynamic Avoidance in Multi-floor Environments
}

\author{Shaomin Liang$^{1}$, Xuanhong Liao$^{2}$, and Shiyao Zhang$^{3}$%
\thanks{$^{1}$Shaomin Liang is with the School of Automation and Intelligent Manufacturing, Southern University of Science and Technology, Shenzhen, China. (e-mail: 12532853@mail.sustech.edu.cn).}
\thanks{$^{2}$Xuanhong Liao is with Guangdong Direct Drive Technology Limited, and also with School of Automation Science and Engineering, South China University of Technology. (e-mail: xuanhong.liao@directdrivetech.com).}%
\thanks{$^{3}$Shiyao Zhang is with the School of Advanced Engineering, Great Bay University, Dongguan City, China. (e-mail: zhangshiyao@gbu.edu.cn).}%
\thanks{\textit{Corresponding author: Shiyao Zhang.}}%
}

\begin{document}

\maketitle
\thispagestyle{empty}
\pagestyle{empty}

\begin{abstract}
In Object Goal Navigation task, existing methods are typically restricted to static and single-floor environments, ignoring cross-floor topologies and dynamic pedestrian, which limits their real-world deployment. To address these limitations, we propose ZONDA, a zero-shot object navigation with dynamic avoidance framework. In particular, ZONDA integrates three core components: (i) Heuristic multi-floor planning: from height-difference traversable maps, enables stair traversal and cross-floor exploration without a platform-specific learned controller; (ii) Multi-view target verification: cross-checks multi-scale observations with a vision-language model, significantly reducing false positives; and (iii) Dynamic pedestrian avoidance: explicitly tracks and predicts moving pedestrians to generate anticipatory behaviors. Evaluated on a real Direct Drive Tech TITA biped robot and extensive simulations on HM3D and MP3D, ZONDA achieves significantly improved results. Moreover, ZONDA can maintain robust navigation on the dynamic benchmark HM3D-DYNA compared to the existing baseline.

\end{abstract}

\section{INTRODUCTION}

Navigating to an unseen object in unknown environments, known as Object Goal Navigation (ObjectNav), is a foundational capability for embodied agents. Although traditional supervised learning methods based on reinforcement learning (RL) \cite{ye2021efficient} have achieved impressive performance in simulated environments, they are typically limited to closed-set object categories and face significant challenges in generalizing to real-world complex environments. As a result, recent studies, e.g. \cite{yokoyama2024vlfm}, have focused on zero-shot ObjectNav. The primary motivation of this paradigm is to overcome the closed-set limitations of traditional learning and eliminate the reliance on expensive task-specific data. By leveraging pretrained foundation models—such as large language models (LLMs) and vision-language models (VLMs)—zero-shot methods equip robots with open-vocabulary semantic reasoning. This enables them to generalize to novel environments and identify unseen object categories without any model fine-tuning.

Despite significant advances in zero-shot ObjectNav, current state-of-the-art frameworks still face three core challenges that limit their use in complex real-world settings.

First, regarding multi-floor exploration, most existing zero-shot methods, e.g. \cite{yokoyama2024vlfm, zhang2025apexnav}, assume a flat, single-floor layout. They collapse 3D semantics into a 2D grid to reduce computational overhead and leverage mature 2D planners. Consequently, this dimensionality reduction leads to failures in multi-floor buildings. While ASCENT \cite{gong2026stairway} tackles the multi-floor challenge using an LLM-driven floor-awareness framework, its low-level execution relies on an RL-based PointNav policy \cite{yokoyama2024vlfm}. This PointNav policy is tightly bound to a specific robot's physical configuration; if the kinematics, chassis size, or camera height changes, the whole policy requires costly retraining. The lack of hardware generalization significantly limits its real-world scalability.

Second, to address target confirmation, existing zero-shot methods, e.g. \cite{yokoyama2024vlfm, zhang2025apexnav, gong2026stairway}, often confirm the target based on just a single view. When the robot is exploring in a quick manner, one glimpse of a visually similar object, shadow, or cluttered background can trigger a false match between a visual distractor and the intended object category. Such false positives can lead to premature termination and task failure.

Third, concerning dynamic environments, most ObjectNav benchmarks, e.g., MP3D \cite{chang2017matterport3d} or HM3D \cite{ramakrishnan2021habitat}, generally assume static environments. In reality, robots shall perform safe navigations around moving pedestrians. Traditional static mapping pipelines treat these dynamic agents as permanent obstacles, leading to navigation failures in dynamic environments. Thus, proactive pedestrian avoidance remains a challenge in zero-shot ObjectNav.

Motivated by these limitations, ZONDA presents a paradigm shift for indoor object navigation. In summary, our main contributions are as follows:
\begin{itemize}
    \item We propose a \textbf{Heuristic Multi-Floor Planner} that removes the need for a platform-specific
  RL-based low-level controller. It enables cross-floor exploration using platform-dependent geometric constraints, rather than retraining a robot-specific navigation policy. Furthermore, it incorporates a \textbf{decoupled dynamic avoidance pipeline} to enable safe and anticipatory navigation in human-shared environments.
    \item We introduce a \textbf{Multi-View Target Verification} module that leverages a vision-language model (VLM) to jointly reason over multi-scale observations, effectively mitigating single-view confirmation bias and reducing false positive matches.
    \item We evaluate the performance of our framework across multiple challenging benchmarks (HM3D, MP3D, and HM3D-DYNA) with extensive comparisons, and demonstrate its efficient real-world deployment on a physical Direct Drive Tech TITA biped robot.
\end{itemize}

\section{RELATED WORKS}

\subsection{Learning-based Object Goal Navigation}
Traditional object navigation methods rely on RL-based methods to train end-to-end policies through extensive simulation interactions. Representative work, e.g. OVRL \cite{yadav2023ovrl}, achieved strong results on benchmarks such as MP3D \cite{chang2017matterport3d} and HM3D \cite{ramakrishnan2021habitat}. However, these methods face two fundamental limitations. First, they are constrained by a closed vocabulary — the agent can only recognize object categories seen during training. Second, the learned policy is closely coupled to the kinematics of a specific robot, so transferring to a different platform requires costly retraining \cite{batra2020objectnav}.

\subsection{Zero-shot Object Goal Navigation}
To overcome the limitations of current learning-based object goal navigation methods, zero-shot ObjectNav turns to pretrained foundation models for open-vocabulary reasoning without task-specific fine-tuning. Early work like ZSON \cite{majumdar2022zson} used CLIP \cite{radford2021learning} to transfer visual-semantic knowledge. VLFM \cite{yokoyama2024vlfm} later improved this by directly mapping the similarity of image and text onto a 2D grid with BLIP-2 \cite{li2023blip}, avoiding information loss from intermediate text and achieving leading performance on single-floor benchmarks. However, its 2D map cannot handle multi-floor environments due to the inability to distinguish overlapping structures across different heights. More recent methods such as ApexNav \cite{zhang2025apexnav} improved exploration efficiency but still operated under the same single-floor assumption.

For multi-floor environments, ASCENT \cite{gong2026stairway} addresses zero-shot navigation across floors. Specifically, it uses multi-level abstraction and coarse-to-fine LLM reasoning to search for targets on different floors. However, it still has two notable shortcomings: its low-level stair navigation still depends on a pre-trained RL policy, so it cannot decouple from a particular robot’s hardware and it does not consider dynamic obstacles.

In this work, we propose ZONDA, unlike prior methods, ZONDA enables cross-floor navigation without a platform-specific learned controller, integrates explicit dynamic pedestrian avoidance, and employs multi-view reasoning for robust target verification. We detail this framework in the below section.

\begin{figure}[t]
    \centering
    \includegraphics[width=\columnwidth]{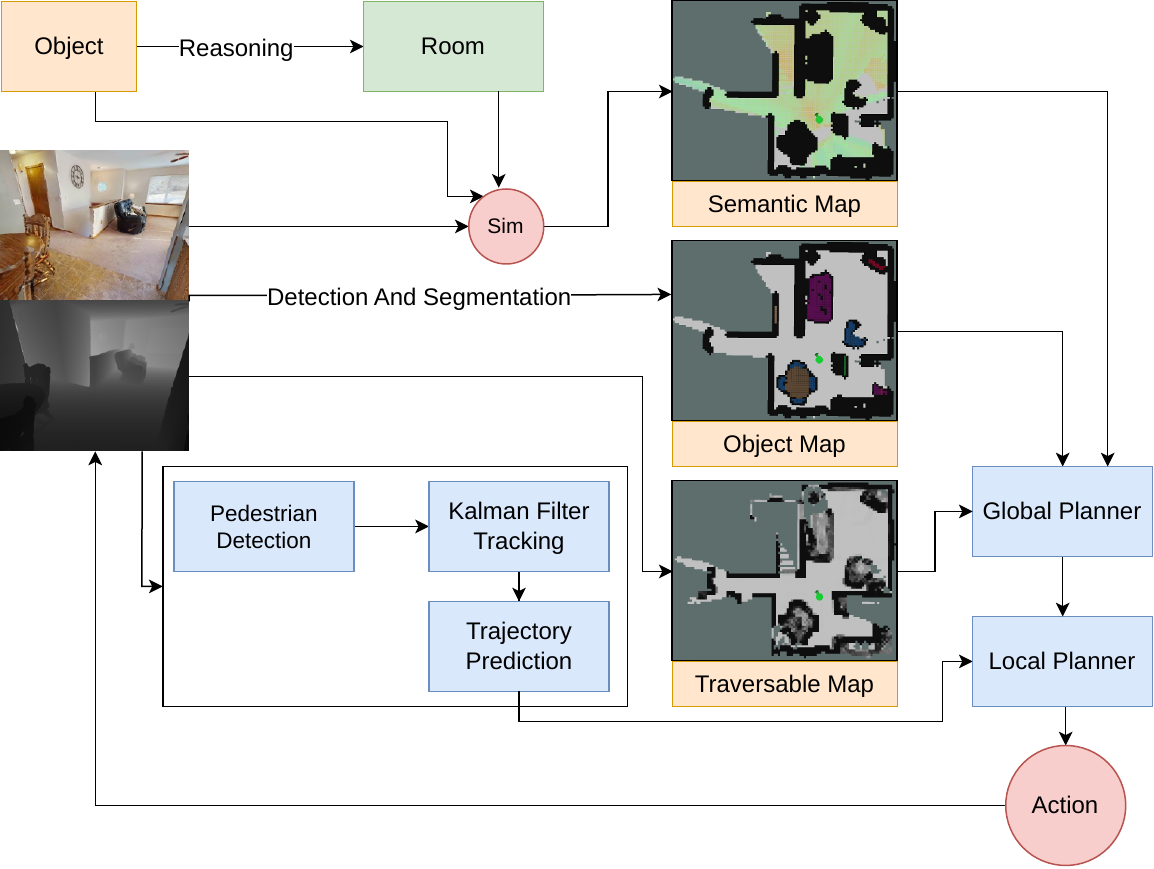}
    \caption{Overview of the ZONDA framework.}
    \label{fig:framework}
\end{figure}

\section{Proposed Methodology}

Fig.~\ref{fig:framework} provides an overview of ZONDA. Built upon an Object-Semantic-Traversability Map that supplies semantic and traversability cues, the framework integrates three core modules: (A) a Heuristic Multi-Floor Planner that handles exploration and cross-floor transitions; (B) a Multi-View Target Verification module that uses a VLM to cross-check candidate targets; and (C) a decoupled pipeline for dynamic pedestrian avoidance. Each module is detailed in the following subsections.

\subsection{Object-Semantic-Traversable Map}

ZONDA maintains three complementary map representations: an object instance map, a semantic heatmap, and a height-difference traversable map, all of which are maintained at a grid resolution of $0.1\,\text{m}$.

\subsubsection{Object Map}
RGB-D images are used to build an object instance map. For each frame, objects are detected and segmented, and their masks are back-projected to 3D using the depth map and camera intrinsics. Each detected object is assigned an incremental confidence score $\delta$:
\begin{equation}
\delta = \text{clip}\bigl(S_{\text{det}} \cdot f_{\text{FOV}} \cdot f_{\text{dist}},\; 0,\; 1\bigr),
\end{equation}
where $S_{\text{det}}$ is the detector confidence score, $f_{\text{FOV}}$ penalizes observations near image edges, and $f_{\text{dist}}$ attenuates scores for objects that are too close or too far. These scores are accumulated into a sparse 3D voxel grid $\mathcal{V}$. To derive the 2D object map, we project the 3D semantic voxel grid onto a bird's-eye view (BEV) plane. Specifically, the semantic label $C(x, y)$ for each 2D cell is determined by the class that maximizes the aggregated semantic confidence along the vertical $z$-axis:
\begin{equation}
C(x,y) = \arg\max_{c} \sum_{z} \mathcal{V}(x,y,z,c).
\end{equation}

\subsubsection{Semantic Map}
Given a target object, an LLM first infers the most likely room type where the object could be found. Both the object name and the inferred room type are embedded into normalized text vectors $\mathbf{v}_{\text{obj}}$ and $\mathbf{v}_{\text{room}}$ using a vision-language model. For each RGB frame, the same model extracts a global image embedding $\mathbf{v}_{\text{img}}$, and the image-level semantic response is computed as the maximum cosine similarity:
\begin{equation}
\text{sim} = \max\bigl(\langle\mathbf{v}_{\text{img}}, \mathbf{v}_{\text{room}}\rangle,\; \langle\mathbf{v}_{\text{img}}, \mathbf{v}_{\text{obj}}\rangle\bigr).
\end{equation}
We project this image-level score onto the map cells $(x,y)$ within the current FOV. To penalize peripheral distortion, we apply a confidence weight $W_{\text{FOV}}(x,y)$ that decays from the center to the edges, yielding $\text{sim}_{\text{w}}(x,y)$. The map is then updated via exponential moving average temporal smoothing:
\begin{equation}
\text{score}_{t}(x,y) = \alpha \cdot \text{score}_{t-1}(x,y) + (1-\alpha) \cdot \text{sim}_{\text{w}}(x,y),
\end{equation}
with $\alpha = 0.9$.

\subsubsection{Traversable Map}
The traversable map is built on a 2D grid spanning the environment. Each cell represents a small ground region and stores the estimated ground height at that location.

To construct a base height map, ZONDA first removes overhead structures (e.g., ceilings) from the point cloud. For each grid cell, after filtering out points above detected vertical gaps, the remaining points are averaged to obtain a local ground height $\bar{h}(x,y)$.

Based on this height map, a height-difference map is computed by comparing each cell's ground height with its 8-connected neighbors:
\begin{equation}
\Delta h(x, y) = \max_{(\Delta x, \Delta y) \neq (0,0)} \bigl| \bar{h}(x, y) - \bar{h}(x+\Delta x, y+\Delta y) \bigr|.
\end{equation}
$H_{\text{agent}}$ denotes the robot's maximum traversable height, which is set according to the physical crossing limits of different robot platforms. Cells with $\Delta h(x,y) < H_{\text{agent}}$ are classified as traversable (e.g., ramps or stairs), while others are treated as vertical obstacles.

\begin{figure}[htbp]
    \centering
    \includegraphics[width=\linewidth]{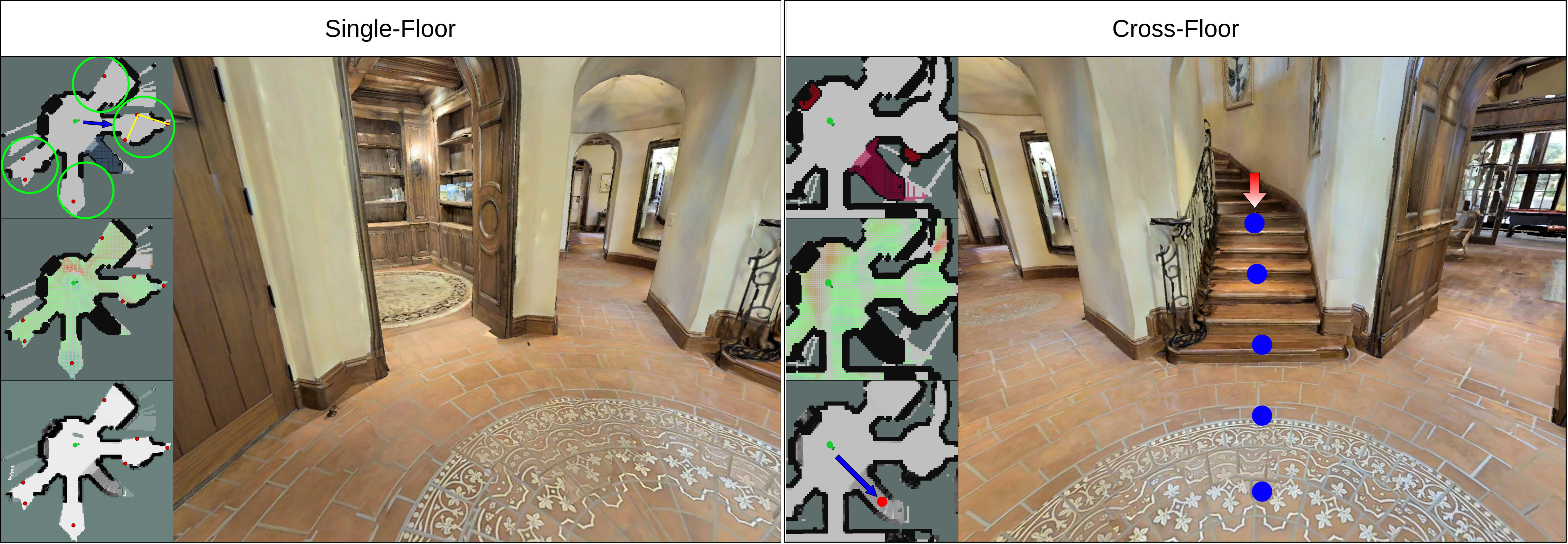}
    \caption{Overview of the exploration planner. \textbf{Left:} Single-floor exploration with frontier clustering and semantic selection. \textbf{Right:} Cross-floor navigation via stair traversability.}
    \label{fig:cross_floor_nav}
\end{figure}

\subsection{Heuristic Multi-Floor Planner}

ZONDA employs a unified heuristic planner for both single-floor exploration and cross-floor transitions (Fig.~\ref{fig:cross_floor_nav}).

During single-floor navigation, safe frontiers are clustered into blocks $\mathcal{B} = \{b_1, \dots, b_M\}$ via DBSCAN on their pairwise geodesic distances. Each block $b_m$ receives a score
\begin{equation}
V(b_m) = w_1 \cdot s_{\max}^{(m)} - w_2 \cdot d_{\text{center}}^{(m)} + \mathcal{R}_{\text{sem}}(b_m),
\end{equation}
where $s_{\max}^{(m)}$ is the maximum semantic score among frontiers within the block, $d_{\text{center}}^{(m)}$ is the geodesic distance from the robot to the block centroid, and $w_1,w_2$ are constant weights. When any frontier in the block exceeds a similarity threshold $\tau_{\text{sem}}$, the reward $\mathcal{R}_{\text{sem}}(b_m)$ is added to further bias the robot towards target-related areas. The best block $b^* = \arg\max V(b_m)$ is selected, and the remaining blocks are queued.

Inside $b^*$, the visitation order of frontiers is formulated as a traveling salesman problem (TSP) to determine an efficient visitation order. A greedy nearest-neighbor heuristic is used: starting from the frontier closest to the robot, the system iteratively visits the nearest unvisited frontier until all are covered.

Besides, when the target is not found and all frontiers on the current floor are completely explored, the planner switches to cross-floor mode. To identify traversable stairways, ZONDA queries its Object-Semantic-Traversable Map. Specifically, a cell is classified as a stair candidate if (i) its semantic label is ``stairs'' with confidence $\delta \geq \delta_{\text{stair}}$, and (ii) it satisfies the traversability condition $\Delta h(x, y) < H_{\text{agent}}$. The robot then plans paths through these verified regions, strictly constraining its movement to traversable cells. Once searching local stair cells are finished, the robot rotates $360^\circ$ and captures a panoramic image for VLM-based scene reasoning. If a stair landing is detected, the search continues on the next flight; if a new floor is confirmed, the previous Object-Semantic-Traversable Map is archived, a new one is initialized, and the connection is registered in an undirected topological graph $\mathcal{G}_{\text{topo}} = (\mathcal{N}, \mathcal{E})$. Here, each edge $e_{i \leftrightarrow j}$ stores the 3D egress and entry positions of the stairs, enabling seamless, bidirectional traversal without re-exploration.

\begin{figure}[htbp]
    \centering
    \includegraphics[width=\linewidth]{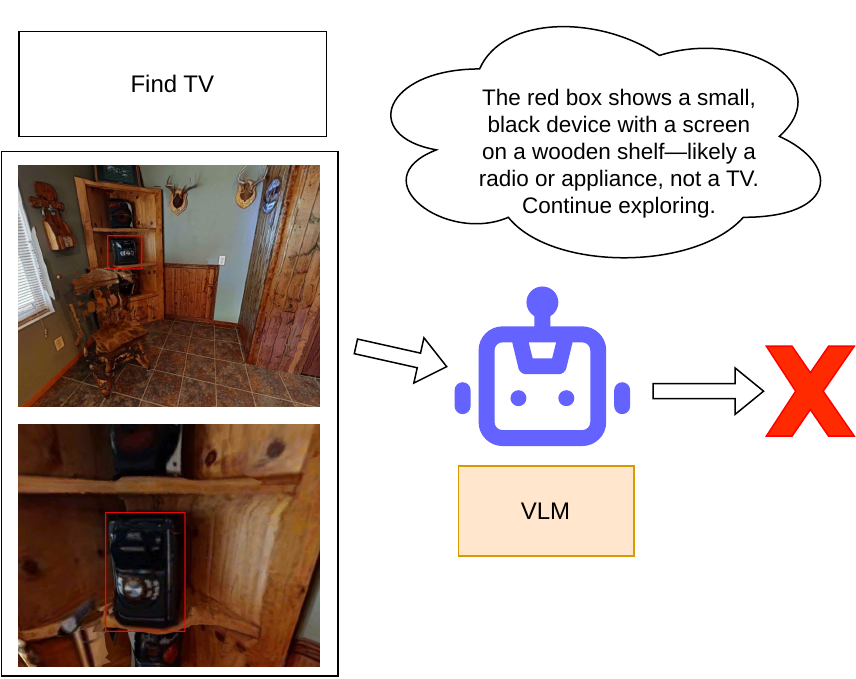}
    \caption{Multi-View Target Verification. Multi-scale observations enable the VLM to combine scene context with detailed appearance, effectively overcoming single-view confirmation bias.}
    \label{fig:multi_view}
\end{figure}

\subsection{Multi-View Target Verification}

To reduce false confirmations caused by single-view ambiguity, ZONDA maintains an observation buffer for each target instance and scores every view by a simple quality metric:
\begin{equation}
Q = S_{\text{raw}} \cdot f_{\text{area}} \cdot f_{\text{edge}},
\end{equation}
where $S_{\text{raw}}$ is the base detection confidence, $f_{\text{area}} = \text{clip}\bigl( \ln(a / a_{\min}) / \ln K,\, 0,\, 1 \bigr)$ favours large, clear observations ($a$ is the target pixel area ratio, $a_{\min}$ the minimum threshold, $K$ a scaling coefficient), and $f_{\text{edge}}$ is a discrete step function that penalizes objects touching the image boundary where features are incomplete.

When the robot reaches a candidate target location, the buffer is queried for the highest-quality views at multiple scales (e.g., near and far fields). As shown in Fig.~\ref{fig:multi_view}, these views are sent together to the VLM for cross-examination. By jointly reasoning about environmental context and close-up detail, the VLM can reject subtle false positives---for instance, distinguishing a small radio from the target TV---substantially improving final confirmation accuracy.

\begin{figure}[htbp]
    \centering
    \includegraphics[width=\linewidth]{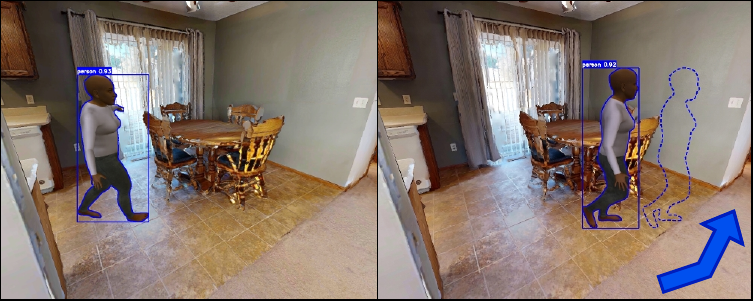}
    \caption{Anticipatory Pedestrian Avoidance. \textbf{Left:} Pedestrian detection and tracking. \textbf{Right:} Predicted future trajectory (dashed outline) and the robot's proactive avoidance behaviour.}
    \label{fig:dynamic_avoidance}
\end{figure}

\subsection{Dynamic Pedestrian Avoidance}

ZONDA handles moving pedestrians through a dedicated pipeline that runs in parallel with static mapping.  
Pedestrians are detected and segmented from RGB-D frames, and their 3D centroids are extracted by back-projecting depth points that fall inside the instance masks.  
These centroids are tracked with a constant-velocity Kalman filter (state $\mathbf{x} = [x, y, v_x, v_y]^\top$), and data association is solved by the Hungarian algorithm with a Euclidean distance gate.

Given the tracked velocity, future positions are extrapolated as $\mathbf{p}(t) = [x + v_x t,\; y + v_y t]^\top$ over a horizon $T_{\text{pred}}=3.0$\,s at 0.5\,s intervals (Fig.~\ref{fig:dynamic_avoidance}).
These predictions enable anticipatory avoidance rather than reactive stopping.

For robust avoidance, both detected pedestrians and their predicted trajectories are treated as obstacles and projected onto the occupancy grid with a 0.50\,m safety inflation. A Heuristic Multi-Floor Planner then selects the next waypoint, followed by A* search for global path planning. For execution, we adopt a dual strategy based on the action space: the robot executes a sequence of discrete commands along the waypoints in discrete environments, while employing an MPPI \cite{williams2016aggressive} local planner to output continuous velocity commands for smooth path tracking in continuous scenarios.

\section{EXPERIMENTS}

\subsection{Experimental Setup}

\noindent\textbf{Datasets and Environments.} We evaluate our method within the Habitat simulator \cite{savva2019habitat} on the standard HM3D and MP3D datasets, both comprising multi-floor residential scenes. HM3D provides 2,000 validation episodes across 20 scenes with 6 COCO object categories. MP3D includes 2,195 validation episodes across 11 scenes with 21 categories, many of which are open-vocabulary classes. To evaluate dynamic obstacle avoidance, we introduce \textbf{HM3D-DYNA}, extending the 2,000 HM3D validation episodes by adding one moving pedestrian to each episode. The pedestrian spawns 3--5\,m from the robot and navigates toward the target along the shortest path at a constant speed of 0.5\,m/s.

\noindent\textbf{Action Space and Execution.} To ensure fair comparisons with prior work, simulation experiments adopt the native discrete action space of Habitat: \{\texttt{MOVE\_FORWARD} ($0.25$\,m), \texttt{TURN\_LEFT} ($30^\circ$), \texttt{TURN\_RIGHT} ($30^\circ$), \texttt{STOP}\}. Instead of exploring directly via these primitive actions, ZONDA plans a global path on the traversable map using A*, converting waypoints into discrete motion sequences.

\noindent\textbf{Evaluation Metrics.} We adopt two standard metrics. Success Rate (SR) is defined as:
\begin{equation}
    \text{SR} = \frac{1}{N}\sum_{i=1}^{N} S_i,
\end{equation}
where $S_i \in \{0,1\}$ indicates if the robot successfully reaches and positively verifies the target within the step budget. Success weighted by Path Length (SPL) measures navigation efficiency:
\begin{equation}
    \text{SPL} = \frac{1}{N}\sum_{i=1}^{N} S_i \cdot \frac{\ell_i}{\max(p_i,\; \ell_i)},
\end{equation}
where $p_i$ is the actual trajectory length and $\ell_i$ is the shortest-path distance. It is worth noting the typical performance bounds for these tasks: achieving an SPL over $30\%$ demonstrates strong navigation efficiency on HM3D. For MP3D, due to its massive scale and highly complex indoor layouts, an SPL above $20\%$ is widely regarded as a highly competitive result.

\noindent\textbf{Implementation Details.} For the perception task, we utilize SegFormer \cite{xie2021segformer} for stair segmentation and SAM2 \cite{ravi2025sam} for mask refinement. Object detection is handled by RT-DETR \cite{zhao2024detrs} (COCO closed-set), Grounding DINO \cite{liu2024grounding} (HM3D open-vocabulary), and OWLv2 \cite{minderer2023scaling} (MP3D open-vocabulary). For vision-language reasoning, Qwen3-VL-Embedding-2B \cite{bai2025qwen3} computes semantic score map embeddings, while Qwen3-VL-Flash \cite{bai2025qwen3} executes the final multi-view target verification. All experiments are conducted on a single NVIDIA RTX 5060 GPU. 

\subsection{Comparison with State-of-the-Art Algorithms}

We comprehensively evaluate ZONDA against state-of-the-art learning-based and zero-shot ObjectNav algorithms. To ensure a clear comparison, we present the results in two distinct settings: static multi-floor navigation (HM3D and MP3D) and dynamic obstacle avoidance (HM3D-DYNA).

\subsubsection{Performance in Static Multi-Floor Environments}

Table \ref{tab:static_results} summarizes the navigation performance on the standard HM3D and MP3D benchmarks. Baseline results in Table~\ref{tab:static_results} are quoted from the original papers with their native perception stacks; thus, the table reports a published system-level comparison rather than a detector- or VLM-controlled reimplementation.

\begin{table}[htbp]
\centering
\caption{Comparison with state-of-the-art methods on static multi-floor benchmarks. Best zero-shot results are highlighted in \textbf{bold}.}
\label{tab:static_results}
\begin{tabular}{l|cc|cc}
\hline
\multirow{2}{*}{Method} & \multicolumn{2}{c|}{HM3D} & \multicolumn{2}{c}{MP3D} \\
                        & SR & SPL  & SR   & SPL  \\
\hline
\multicolumn{5}{c}{\textit{Learning-based}} \\
\hline
PONI\cite{ramakrishnan2022poni}                    & --  & --   & 31.8 & 12.1 \\
SemExp\cite{chaplot2020object}                  & --  & --   & 36.0 & 14.4 \\
PIRLNav\cite{ramrakhya2023pirlnav}                 & 64.1& 27.1 & --   & --   \\
\hline
\multicolumn{5}{c}{\textit{Zero-shot}} \\
\hline
VLFM\cite{yokoyama2024vlfm}                    & 52.5& 30.4 & 36.4 & 17.5 \\
SG-Nav\cite{yin2024sg}                  & 54.0& 24.9 & 40.2 & 16.0 \\
OpenFMNav\cite{kuang2024openfmnav}               & 54.9& 24.4 & 37.2 & 15.7 \\
APEXNav\cite{zhang2025apexnav}                 & 59.6& 33.0 & 39.2 & 17.8 \\
ASCENT\cite{gong2026stairway}                  & 65.4& \textbf{33.5} & 44.5 & 15.5 \\
\hline
\textbf{ZONDA (ours)}   & \textbf{66.5} & 33.0 & \textbf{48.2} & \textbf{21.5} \\
\hline
\end{tabular}
\end{table}

On HM3D, ZONDA achieves an impressive success rate (SR) of 66.5\%, outperforming both the strongest zero-shot baseline, ASCENT (65.4\%), and the learning-based PIRLNav (64.1\%). While ASCENT holds a marginal edge in SPL (33.5\% vs. 33.0\%), ZONDA's superior SR clearly demonstrates the efficacy of our multi-view target verification module in filtering out false positives.

On the expansive, open-vocabulary MP3D dataset, ZONDA sets a new state-of-the-art with 48.2\% SR and 21.5\% SPL, outperforming ASCENT by 3.7 and 6.0 percentage points,respectively. This remarkable navigational efficiency demonstrates that our heuristic cross-floor strategy and multi-view confirmation successfully minimize redundant exploration across massive indoor environments.

\begin{table}[htbp]
\centering
\caption{Evaluation on the proposed HM3D-DYNA benchmark.}
\label{tab:dynamic_results}
\begin{tabular}{l|cc}
\hline
\multirow{2}{*}{Method} & \multicolumn{2}{c}{HM3D-DYNA} \\
                        & SR & SPL  \\
\hline
ASCENT                  & 30.9 & 16.6 \\
\textbf{ZONDA (ours)}   & \textbf{48.8} & \textbf{24.7} \\
\hline
\end{tabular}
\end{table}

\subsubsection{Performance in Dynamic Environments}

To evaluate the robustness of our pedestrian avoidance pipeline, we compare ZONDA against ASCENT on the proposed HM3D-DYNA benchmark, where moving pedestrians actively interfere with the navigation path.

As shown in Table~\ref{tab:dynamic_results}, dynamic obstacles severely degrade the performance of methods without explicit avoidance. ASCENT lacks a dedicated dynamic perception module, and its success rate drops to 30.9\%, as the robot is frequently blocked by or collides with moving pedestrians. In contrast, ZONDA maintains a 48.8\% SR and achieves 24.7\% SPL. These results demonstrate that by explicitly tracking and predicting pedestrian motions, ZONDA enables safe and efficient navigation even in shared human-robot spaces.

\subsection{Ablation Studies}

In this part, we conduct ablation experiments to quantify the contribution of three key components: the heuristic block-based exploration strategy, the cross-floor navigation module, and the multi-view target verification module. Table~\ref{tab:ablation} reports the results.

\begin{table}[htbp]
\centering
\caption{Ablation results on HM3D and MP3D.}
\label{tab:ablation}
\begin{tabular}{l|cc|cc}
\hline
\multirow{2}{*}{Method} & \multicolumn{2}{c|}{HM3D} & \multicolumn{2}{c}{MP3D} \\
                        & SR (\%) & SPL (\%) & SR (\%) & SPL (\%) \\
\hline
\textbf{ZONDA (full)}   & \textbf{66.5} & \textbf{33.0} & \textbf{48.2} & \textbf{21.5} \\
\hline
w/o heuristic blocks    & 62.6 & 26.8 & 38.6 & 13.4 \\
w/o cross-floor         & 57.8 & 28.5 & 44.2 & 20.7 \\
w/o multi-view verify   & 41.5 & 23.4 & 30.2 & 11.6 \\
\hline
\end{tabular}
\end{table}

\textbf{Heuristic block-based exploration.}
In this variant, frontier clustering and block scoring are disabled (\textit{w/o heuristic blocks}); the robot simply navigates to the nearest unexplored frontier. On HM3D, SR drops from 66.5\% to 62.6\%, and SPL from 33.0\% to 26.8\%; on MP3D, SR falls from 48.2\% to 38.6\%, and SPL from 21.5\% to 13.4\%. The consistent decline confirms that grouping frontiers into blocks and selecting them with semantic rewards is essential for efficient exploration in multi-floor environments.

\textbf{Cross-floor navigation.}
When the robot is restricted to the starting floor (\textit{w/o cross-floor}), HM3D SR falls from 66.5\% to 57.8\%, SPL from 33.0\% to 28.5\%. MP3D SR drops from 48.2\% to 44.2\%, SPL from 21.5\% to 20.7\%. The smaller MP3D gap reflects that more targets there are already on the start floor, but both benchmarks still benefit from cross-floor transitions.

\textbf{Multi-view target verification.}
When multi-view verification is removed and the robot relies only on the first detection view (\textit{w/o multi-view verify}), HM3D SR drops from 66.5\% to 41.5\%, SPL from 33.0\% to 23.4\%; MP3D SR falls from 48.2\% to 30.2\%, SPL from 21.5\% to 11.6\%. This decline indicates that a single observation is insufficient for reliable target verification, whereas aggregating multiple discriminative views improves robustness.

\begin{figure}[htbp]
    \centering
    \includegraphics[width=\linewidth]{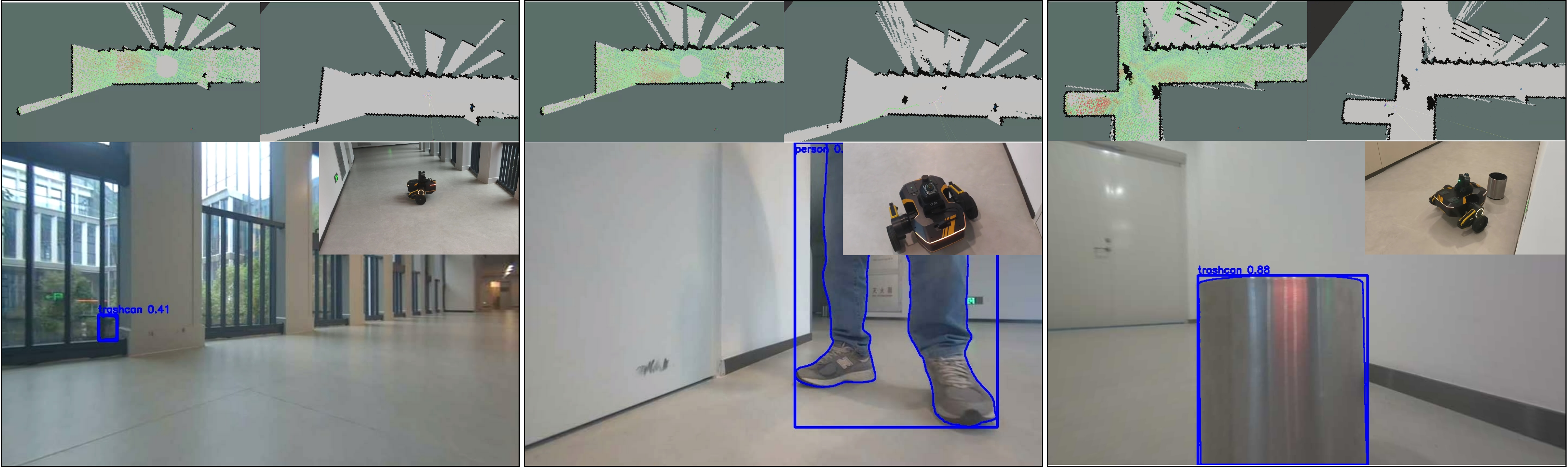}
    \caption{Real-world Assessment on the Direct Drive Tech TITA.}
    \label{fig:real}
\end{figure}

\subsection{Real-World Deployment}

We deploy ZONDA using an off-board computing architecture. Heavy model inference is executed on a local workstation equipped with an Intel Core i7-14700KF CPU and an NVIDIA RTX 5060 Ti GPU, while physical navigation is carried out by a Direct Drive Tech TITA wheel-legged biped robot. The software stack is built on ROS 2, enabling seamless network communication between the local compute node and the robot's sensors. In contrast to the discrete action execution used in simulation, the physical robot requires smooth continuous motion. Therefore, we employ MPPI as a low-level local planner to execute actions, continuously tracking the generated waypoints while proactively avoiding dynamic pedestrians.

To verify the real-world performance of our system, we deploy the robot in a dynamic indoor office scenario. This evaluation serves to test the model's continuous control and spatial reasoning in a physical setting. As illustrated in Fig.~\ref{fig:real}, the system maintains a semantic map for global exploration and an object map for target localization (the traversable map is omitted for this single-floor setup). All non-platform-specific parameters are identical to those used in the HM3D benchmark. Platform-dependent parameters, including $H_{\text{agent}}$ and the safety inflation radius, are set according to the physical specifications of the TITA robot. Fig.~\ref{fig:real} shows a representative successful episode: the robot effectively avoids a moving pedestrian, accurately localizes the ``trash can'', and cleanly terminates the task.

\section{CONCLUSION}
In this paper, we introduce ZONDA, a zero-shot ObjectNav framework jointly handling multi-floor navigation, robust target verification, and dynamic pedestrian avoidance without task-specific training. Specifically, the Heuristic Multi-Floor Planner enables stair traversal without relying on a platform-specific RL controller and efficient cross-floor navigation, the Multi-View Target Verification module mitigates single-view false positives, and the Dynamic Pedestrian Avoidance pipeline provides anticipatory behaviors in human-shared environments. Experimental studies demonstrate that by integrating these three components, ZONDA not only achieves significant performance on static HM3D and MP3D benchmarks, but also exhibits exceptional robustness on our dynamic HM3D-DYNA dataset where ZONDA substantially outperforms ASCENT. Ablation studies verify the contribution of each module, and real-world deployment confirms its transferability. Future work will extend the framework to a wider range of dynamic obstacles and outdoor environments.



\bibliographystyle{IEEEtranBST/IEEEtrans}
\bibliography{IEEEtranBST/ref} 

\end{document}